\documentclass{article}
\pdfoutput=1

\usepackage{arxiv}
\usepackage{bm}
\usepackage{amsmath}
\usepackage{graphicx}
\usepackage{subfig}
\usepackage{amssymb}
\usepackage{float}
\DeclareMathOperator{\sign}{sign}

\title{Learning Walk and Trot from the Same Objective Using Different Types of Exploration}

\author{
Zinan Liu$^{*\dagger}$, Kai Ploeger\thanks{Both authors contributed to this work equally,
        \{kai.ploeger, zinan.liu\}@stud.tu-darmstadt.de} $^{\,\dagger}$, 
Svenja Stark$^{\dagger}$, Elmar Rueckert\thanks{Technische Universit\"at Darmstadt, FB Informatik, FG Intelligent Autonomous Systems, Darmstadt, Germany} and Jan Peters$\dagger$\thanks{Max Planck Institute for Intelligent Systems, Department of Empirical Inference, Tübingen, Germany\newline
This project has received funding from the European Union’s Horizon 2020 research and innovation programmes under grant agreement No 713010 (GOAL-Robots) and 640554 (SKILLS4ROBOTS).
}}

\makeatletter
 \def\@textbottom{\vskip \z@ \@plus 1pt}
 \let\@texttop\relax
\makeatother
\begin{document}
\maketitle

\begin{abstract}
In quadruped gait learning, policy search methods that scale high dimensional continuous action spaces are commonly used. In most approaches, it is necessary to introduce prior knowledge on the gaits to limit the highly non-convex search space of the policies. In this work, we propose a new approach to encode the symmetry properties of the desired gaits, on the initial covariance of the Gaussian search distribution, allowing for strategic exploration. Using episode-based \textit{likelihood ratio policy gradient } and \textit{relative entropy policy search}, we learned the gaits walk and trot on a simulated quadruped. Comparing these gaits to random gaits learned by initialized diagonal covariance matrix, we show that the performance can be significantly enhanced.
\end{abstract}

\section{Introduction}
In nature, animals have developed extensive \textit{gaits} to adapt to the different terrestrial terrain and situations, such as a horse galloping for faster speed, or a lizard trotting for a stable locomotion.  In recent years, quadrupedal gait learning has attracted some research interest in robotics. Quadruped gaits offer a wide range of different movement patterns. As the cyclic movements of all four legs are similar, the gaits can be 
categorized mainly by the timing and order of the footfall, which can be represented as phase gaps among the trajectories of each leg.

In the presented work we learn open loop control policies for various gaits, focusing on \textit{walk} and \textit{trot}. In walk the leg trajectories are separated by quarter-phase gaps, resulting in an equidistant footfall, whereas in trot diagonal pairs of legs move synchronously and are separated by half-phase gaps. Other gaits that can be learned using the described approach are \textit{bound} and \textit{pace}. We show how these symmetry properties can be encoded in the parameter space of the chosen policy representation, in order to enhance the initial exploration and reliably learn the chosen gaits.

Neither do we fully define the gait in the policy representation as in \cite{Elma,kohl2004policy,saggar2006autonomous,chernova2004evolutionary}, nor do we learn random gaits~\cite{DeepmindRich2017} which could lead to a highly non convex problem. We only initialize the exploration. This way we can learn different gaits as a set of skills sharing a common representation, thus allowing for choosing gaits suitable to the current terrain. 

Traditional reinforcement learning methods, such as \textit{TD-Learning}, typically estimate the expected long-term reward at each state $\boldsymbol{s}$ and each time step $t$, to evaluate the quality of executing an action $\boldsymbol{u}$ in state $\boldsymbol{s}$. This is referred as the \textit{value function} $V^{\pi}_t(\boldsymbol{s})$. Given a certain state and all possible actions, the \textit{action-value function} $(\boldsymbol{s}_t, \boldsymbol{u}_t)$ is computed, an optimal action $\boldsymbol{u}^*_t$is then selected to optimize the action-value function \cite{sutton}. However, this approach would be problematic in high dimensional state action space, since the whole space would need to be exhaustively covered. In addition, gait learning of a quadruped would also require continuous states and actions, meaning we would have to resort to value function approximation. 

On the other hand, \textit{Policy Search} methods offer the possibility to scale a high dimensional continuous action space to a reduced search space of possible policies. This can be done by parameterizing the \textit{search distribution} $p{(\boldsymbol{\theta})}$, and directly operate and update in the parameter space $\boldsymbol{\Theta}$, $\boldsymbol{\theta} \in \boldsymbol{\Theta}$ \cite{survey}. In this work, we learn a Gaussian search distribution $p(\boldsymbol{\theta})= \mathcal{N}(\boldsymbol{\theta}|\boldsymbol{\mu, \Sigma})$ that maximizes the expected reward, using \textit{Likelihood Ratio Policy Gradient} \cite{PG} and \textit{Relative Entropy Policy Search} \cite{REPS}. The trajectories of the quadruped are computed by a linear combination of the von Mises basis functions $\boldsymbol{b}$ and their corresponding weights $\boldsymbol{w}$, i.e., $\tau = \boldsymbol{b}^T\boldsymbol{w}$. The generated trajectories are fed to a PD controller to deterministically compute the motor torques.

\section{Related Works}
This section presents first the policy representation to 
be used in the experiments in conjunction with the subsequently
presented episode-based policy search algorithms.

\subsection{Policy Representation}
The explored policies are central pattern generators using a linear representation $\boldsymbol{q}_{des}=\boldsymbol{b}^{T}\boldsymbol{w}$, $\boldsymbol{b}$ being the von Mises basis functions,
\begin{align}
	b_i(z)=\exp{\left(\frac{\cos(2\pi(z_t-c_i))}{h}\right)},
	\label{VonM}
\end{align}
where $h$ is the width, $c_i$ are the equidistant centers of basis function, $z(t) = \delta_zt$ is the phase, which is linear in time $t$ with a temporal scaling factor $\delta_z$ describing the speed of executing the trajectory \cite{Elma,ProMP_NIPS}. The policy parameters $\boldsymbol{\theta}$ to be learned then consist of the weights $\boldsymbol{w}$, and the temporal scaling factor $\delta_z$.

In order to simplify the learning problem, we hand tuned the initial standing posture of the quadruped to prevent it from falling. The learning exploration of each rollout begins then with the standing initialization.

The trajectories $\boldsymbol{q}_{\mathrm{des}}=\boldsymbol{\tau({\boldsymbol{\theta}})}$ defined by the learned policy, are then fed to a PD controller to deterministically compute the motor torques $\boldsymbol{u}$,
\begin{equation}
\boldsymbol{u}=K_p(\boldsymbol{q}_{\mathrm{des}}-\boldsymbol{q})+K_D(\boldsymbol{\dot{q}}_{\mathrm{des}}-\boldsymbol{\dot{q}}),\label{PD}
\end{equation}
where the desired velocities $\boldsymbol{\dot{q}}_{\mathrm{des}}$ are approximated using forward differences of $\boldsymbol{q}_{des}$, and $\boldsymbol{q}$, $\boldsymbol{\dot{q}}$ are the observed state variables.

\subsection{Episode-based Policy Search Algorithms}
 
Episode-based policy search algorithms typically maximize the expected reward ${J_{\boldsymbol{\mu},\boldsymbol{\Sigma}}} = \mathbb{E}_{\boldsymbol{\theta}}[R(\boldsymbol{\theta})]$ over a Gaussian search distribution $\mathcal{N}(\boldsymbol{\theta}|\boldsymbol{\mu},\boldsymbol{\Sigma})$, throughout the episode, w.r.t the distribution means $\boldsymbol{\mu}$~\cite{survey}. Independent of the policy search algorithm, the parameter vector $\boldsymbol{\theta}^*$, used for execution after a terminated learning process, is chosen as the final update of $\boldsymbol{\mu}$.
 

\subsubsection{Likelihood Ratio Policy Gradient}

Policy Gradient methods~\cite{PGjan} use gradient ascent to maximize the expected return ${J}_{{\boldsymbol{\mu}},\boldsymbol{\Sigma}}$ over the search distribution w.r.t. the parameters ${\boldsymbol{\mu}}$,
\begin{align}
	{\boldsymbol{\mu'}} = {\boldsymbol{\mu}} + \alpha\nabla_{{\boldsymbol{\mu}}}J_{{\boldsymbol{\mu}},\boldsymbol{\Sigma}}
\end{align}
In the following we estimate the gradient according to the Likelihood Ratio Policy Gradient method (LRPG) \cite{PG} with the batch mean reward $\bar{R}$ as baseline, sampling from a Gaussian search distribution $\mathcal{N}\boldsymbol{\theta}_i|\boldsymbol{\mu},\boldsymbol{\Sigma})$. A derivation can be found in~\ref{apdx:PG}.
\begin{align}
	\nabla_{{\boldsymbol{\mu}}} J_{{\boldsymbol{\mu}},\boldsymbol{\Sigma}} &= \mathbb{E}_{\boldsymbol{\tau}}\{\nabla_{{\boldsymbol{\mu}}}\log P(\boldsymbol{\tau}|\boldsymbol{\mu},\boldsymbol{\Sigma})R(\boldsymbol{\tau})\}
    \label{eq:likelihood_ratio}\\
	&\approx \frac{1}{N}\sum_{i}^N \boldsymbol{\Sigma}^{-1}(\boldsymbol{\theta}_i-{\boldsymbol{\mu}})(R(\boldsymbol{\theta}_i)-\bar{R})
\end{align}

\subsubsection{Relative Entropy Policy Search}

Similar to the Expectation Maximization (EM) algorithm \cite{EM}, Relative Entropy Policy Search (REPS) also updates the mean and covariance of the search distribution to maximize the expected reward $J_{\boldsymbol{\mu, \Sigma}}$. In addition, REPS introduces the Kullback-Leibler divergence, as to upper bound the parameter update of the search distribution, thus forcing the parameter update to stay close to the sampled data. Considering the old distribution $q(\boldsymbol{\theta|\mu, \Sigma})$, and the newly estimated distribution $p(\boldsymbol{\theta|\mu', \Sigma'})$, the KL divergence is bounded by $\epsilon$, where $\text{KL}(q(\boldsymbol{\theta|\mu, \Sigma})|p(\boldsymbol{\theta|\mu', \Sigma'})) \leq \epsilon$. This then resolves to a constraint optimization problem 

\begin{align}
\begin{split}
\max_{\boldsymbol{\mu', \Sigma'}} J_{\boldsymbol{\mu', \Sigma'}} = \max & \int_{\boldsymbol{\Theta}} p(\boldsymbol{\theta|\mu', \Sigma'})R(\boldsymbol{\theta})d\boldsymbol{\theta},\\
\text{s.t.}\quad \epsilon & \geq \int_{\boldsymbol{\Theta}}  p(\boldsymbol{\theta|\mu', \Sigma'})\log \frac{p(\boldsymbol{\theta|\mu', \Sigma'})}{q(\boldsymbol{\theta|\mu, \Sigma})},\\
1 & = \int_{\boldsymbol{\Theta}} p(\boldsymbol{\theta|\mu', \Sigma'}) d\boldsymbol{\theta}.
\end{split}
\end{align}

The optimization can be solved by using the method of Lagrangian multipliers, resulting in a closed form solution of the new distribution
\begin{align}
p(\boldsymbol{\theta|\mu', \Sigma'}) \propto q(\boldsymbol{\theta|\mu, \Sigma}) \exp{\left(\frac{R(\boldsymbol{\theta})}{\eta}\right)},
\end{align}

where $\eta$ is obtained by minimizing the dual function $g(\eta)$,  approximated by samples,
\begin{align}
g(\eta) \approx \epsilon\eta+\eta\log\frac{1}{N}\sum_{i}q(\boldsymbol{\theta_i}|\mu, \Sigma)
\exp{\left(\frac{R(\boldsymbol{\theta_i})}{\eta}\right)}
\end{align}

The new distribution is then estimated by fitting a parametric distribution $p(\boldsymbol{\theta}|\boldsymbol{\mu}', \boldsymbol{\Sigma'})$ to the samples $\boldsymbol{\theta_i}$ and the corresponding reward $R(\boldsymbol{\theta_i})$. The parametric distribution is estimated using weighted maximum likelihood, where the weightings $d_i=\exp\left(\frac{R(\boldsymbol(\theta_i))}{\eta}\right)$

Given the Gaussian distribution of our set up, $p(\boldsymbol{\theta|\mu, \Sigma})$, the weighted maximum likelihood solution for updating the distribution, namely $\boldsymbol{\mu'}$ and $\boldsymbol{\Sigma'}$, is given by \cite{EM}

\begin{align}
\boldsymbol{\mu'}=\frac{\sum_{i=1}^{N}d_i\boldsymbol{\mu_i}}{\sum_{i=1}^{N}d_i}, \quad \boldsymbol{\Sigma'}=\frac{\sum_{i=1}^N d_i(\boldsymbol{\mu_i-\boldsymbol{\mu}})(\mu_i-\boldsymbol{\mu}^{T})}{Z},
\end{align}

where
$$
Z = \frac{(\sum_{i=1}^{N}d_i)^2 - \sum_{i=1}^N(d_i)^2}{\sum_{i=1}^N d_i}.
$$

\section{Evaluations}
In this section, we evaluate the performance of the Likelihood Ratio Policy Gradient and the Relative Entropy Policy Search algorithm given different initializations for covariance of the Gaussian search distribution, which cause specific desired gaits to be learned.

\subsection{Simulated Experimental Setup}

We used the OpenAI Gym toolkit~\cite{gym} with the MuJoCo simulator~\cite{mujoco}, and adapted the predefined \textit{ant} environment to a \textit{quadruped} with eight parallel joints. The quadruped is a complete symmetric robot, with four hip joints and four knee joints, where each joint is driven by an actuator, rendering in total eight degrees of freedom. Four basis functions are used for each joint, allowing for quarter-phase gaps, while reducing the amount of weights to learn. In total, the parameters to be learned consist of 32 weights $\boldsymbol{w}$, and the temporal scaling factor $\delta_z$. 

Throughout the simulation, a batch size of 50 is used, and each learning episode consists of 200 time frames, equivalent to 2 seconds given that the time step is 0.01s. The number of policy updates is fixed to 60. 
10 trials are carried out for each configuration to evaluate the robustness of the algorithms.
\subsection{Reward Function and Algorithm Comparison}

We devise the \textit{reward function} consisting of the forward translation, costs of the robot falling over $c_{\mathrm{fall}}(\boldsymbol{\tau})$, contact costs between the feet and the ground $c_{\mathrm{ct}}(\boldsymbol{\tau})$, and the control costs for the motors $c_{\mathrm{ctrl}}(\boldsymbol{\tau})$
 \begin{align}
 R_{\tau} = \frac{1}{T}(\sign({y})\sqrt[]{x^2+y^2}- c_{\mathrm{ctrl}}(\boldsymbol{\tau}) - c_{\mathrm{ct}}(\boldsymbol{\tau}))-c_{\mathrm{fall}}(\tau),
 \end{align}
where $x$ and $y$ denotes the total distance traveled in x and y direction in one episode, and $T$ is the duration of one episode. Since each learning process is initialized with a standing position, only one walking direction $\sign({y})$ is rewarded, which is the forward direction.

\begin{figure}[ht]
  \centering
    \includegraphics[width=0.5\textwidth]{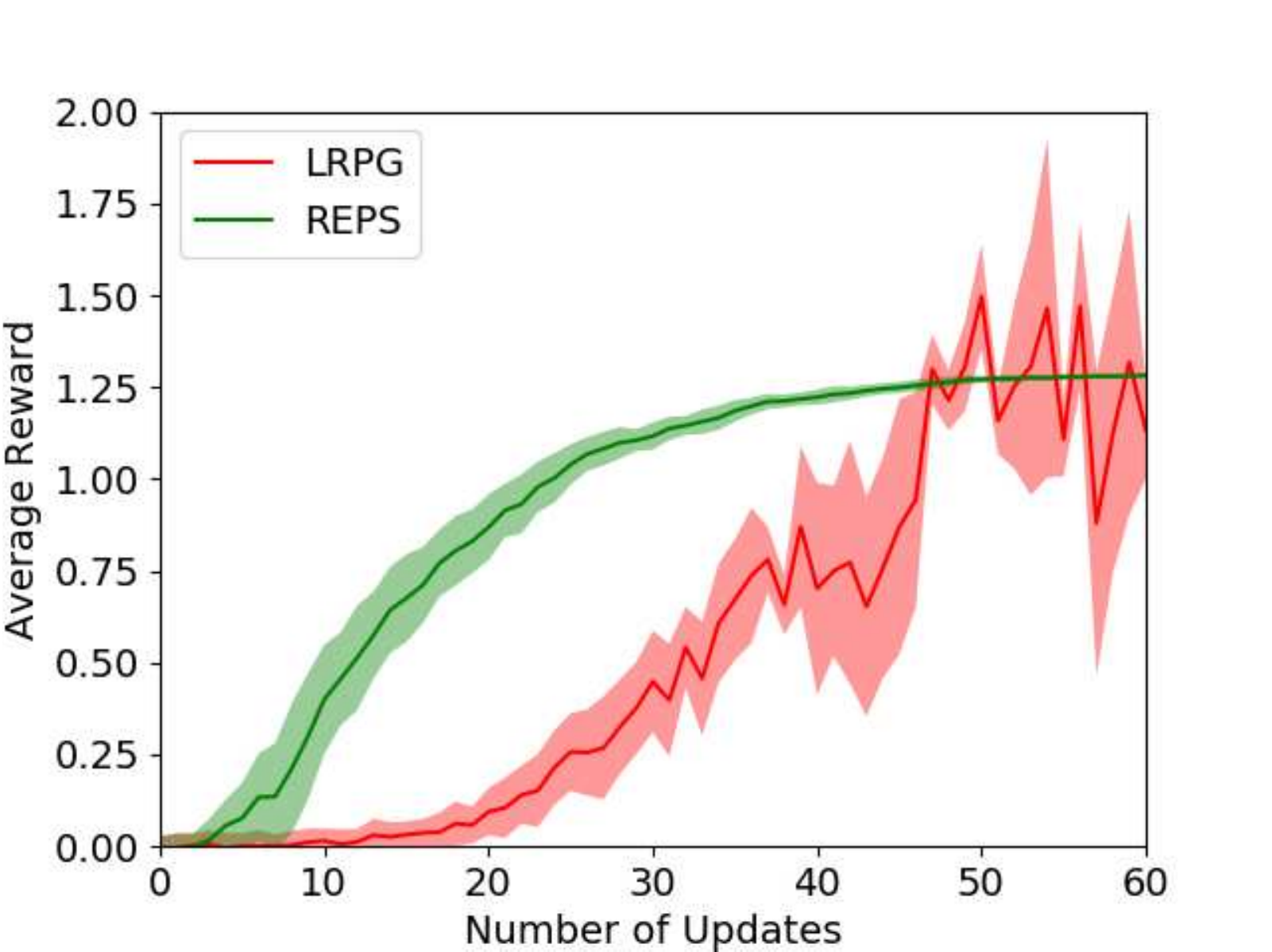}
  \caption{Typical learning curves of Likelihood Ratio Policy Gradient (LRPG) and Relative Entropy Policy Search (REPS) with diagonal covariance initialization, averaged over 50 rollouts per update.}
  \label{diag compare}
\end{figure}

To compare the learning behavior of REPS and LRPG, both are evaluated with hand tuned hyper parameters, and diagonal covariance initializations $\boldsymbol{\Sigma} = \sigma^2 \boldsymbol{I}$. A learning rate of $\alpha_{LRPG} = 0.001$ is used for LRPG, and a KL bound of $\epsilon_{KL} = 0.11$ for REPS. The learning curves of the expected reward and corresponding standard deviation are shown in Figure~\ref{diag compare}, where REPS learns relatively fast in the initial phase before converging to a stable solution. This is due to the applied KL bound, forcing the update to stay close to the old distribution~\cite{REPS}. However, REPS also suffers from premature convergence, since the updates of the covariance $\boldsymbol{\Sigma}$ reduce further exploration~\cite{abdolmaleki2015model}. LRPG shows poor convergence properties, yet the exploration sometimes outperforms policies fond with REPS. 

\subsection{Different Initializations}

Aside from a diagonal initialization of the search distributions covariance matrix, a customized covariance matrix can be used to incorporate prior knowledge about the symmetry properties of a specific gait,
\begin{equation}
	\boldsymbol{\Sigma} = \sigma^2 (\boldsymbol{I} + \gamma \boldsymbol{O}),
    \label{eq:split_cov}
\end{equation}
where $\sigma^2$ is the variance along each dimension, $\boldsymbol{O}$ is a symmetric matrix with elements $p_{ij} \in \{0;1\}$, and $\gamma \in (0,1)$ is the degree of coupling. This way the exploration is limited in certain search directions of the parameter space $\boldsymbol{\Theta}$, promoting the learning of movements with desired symmetry properties.

\begin{figure}[ht]
  \centering
  \subfloat[Learning curves averaged over 10 trials of Likelihood Ratio Policy Gradient, each using specific covariance initializations at a batch-size of 50]{\includegraphics[width=0.49\textwidth]{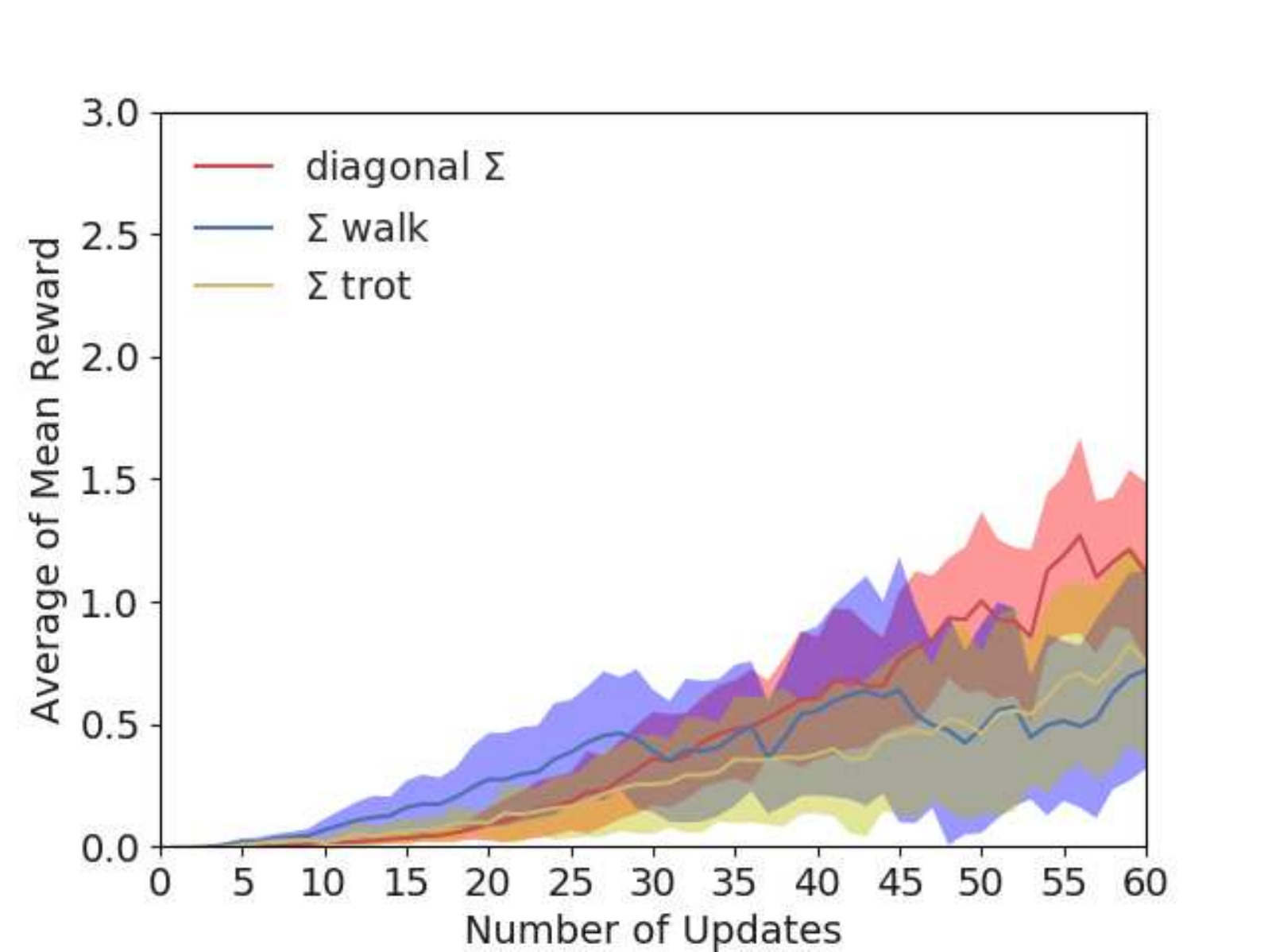}\label{extendedPG}}
  \hfill
  \subfloat[Learning curves averaged over 10 trials of Relative Entropy Policy Search, each using specific covariance initializations at a batch-size of 50]{\includegraphics[width=0.49\textwidth]{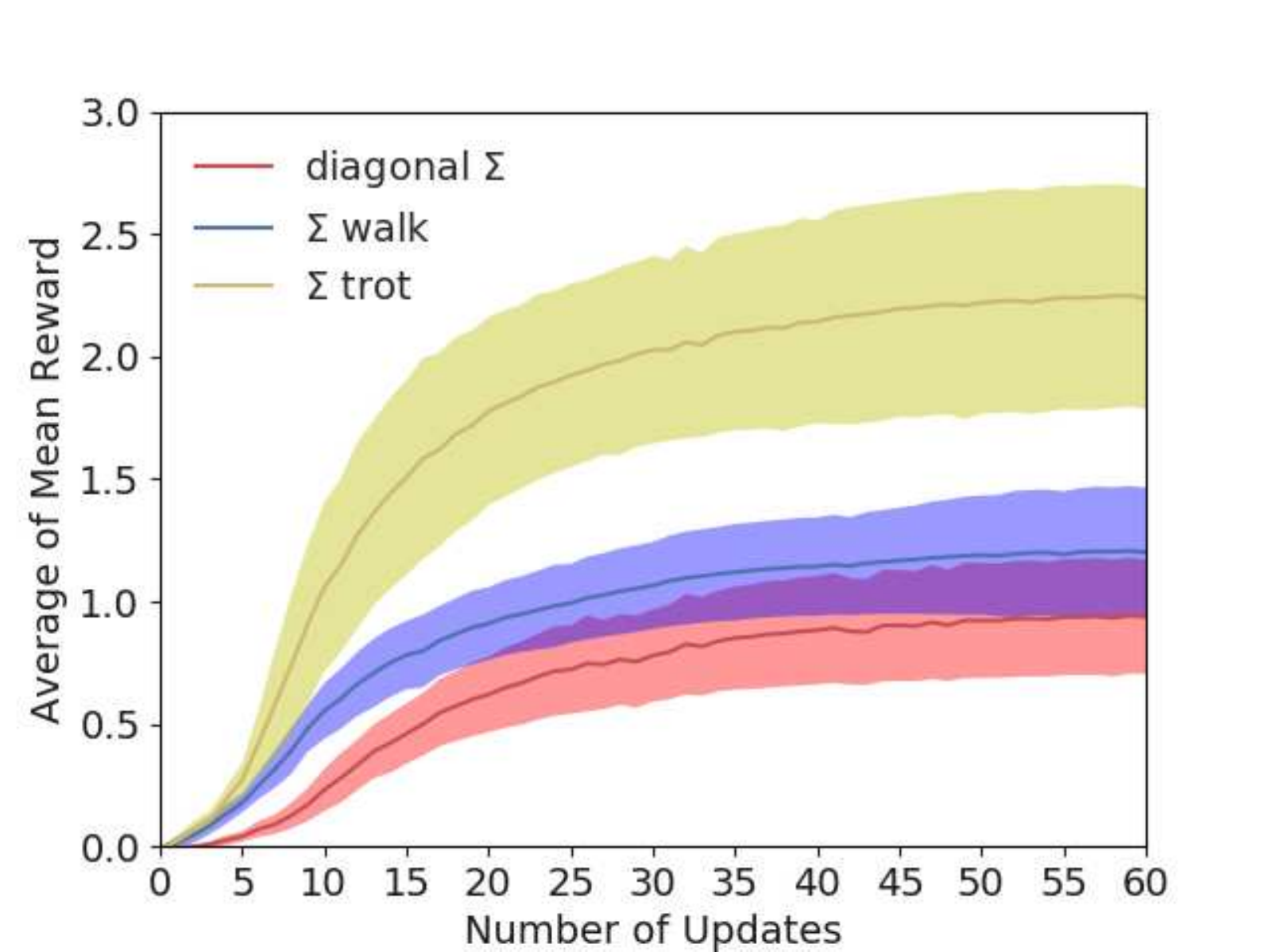}\label{extendedREPS}}
  \hfill
    \caption{Snapshots of different gaits learned via full covariance initialization.}
  \label{extended}
\end{figure}

In Figure~\ref{extendedPG} and \ref{extendedREPS} three different initializations are applied: a diagonal and two non-diagonal matrices specific to walk and trot, where, the degree of coupling is $\gamma = 0.9$ Each learning curve is averaged over 10 trials to evaluate the robustness of the policy updates, and the colored regions indicate the standard deviation of the expected reward over the 10 trials. 
In Figure~\ref{extendedREPS} we can see that using the non-diagonal initializations, the slope of the learning curves as well as the expected reward of the learned policy increase compared to the diagonal initialization, indicating that the KL bounded updates of REPS are able to exploit the prior knowledge, even though it changes the covariance matrix in each update. In addition it shows that for this robot the trot gait is inherently faster than the walk gait, which is the usual case for most quadrupeds~\cite{xi2016selecting}
In Figure~\ref{extendedPG} however, LRPG tends to show decreased performance for non-diagonal initializations. 

\begin{figure}[h]
  \centering
  \subfloat[Snapshots of the walking gait]{\includegraphics[width=1\textwidth]{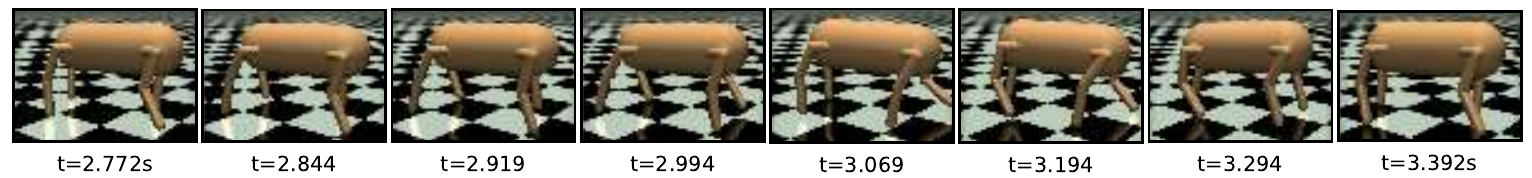}\label{walk}}
  \hfill
  \subfloat[Snapshots of the trotting gait]{\includegraphics[width=1\textwidth]{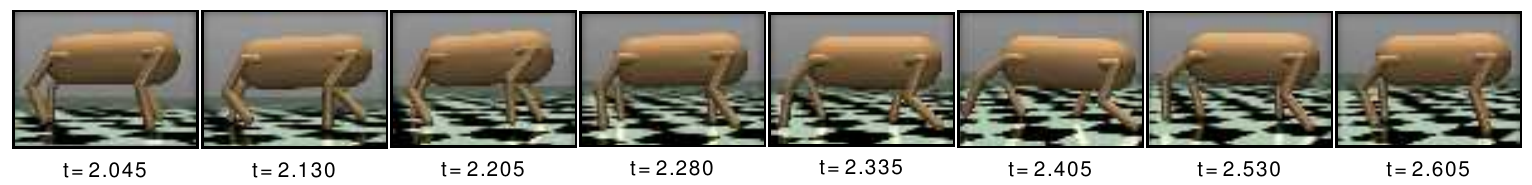}\label{trot}}
  \hfill
    \caption{Snapshots of different gaits learned via full covariance initialization.}
  \label{snapshots}
\end{figure}

Using LRPG as well as REPS, movements matching to the desired gaits are reliably learned given a degree of coupling  for $\gamma > 0.7$. In Figure~\ref{snapshots}, the snapshots of the learned walk and trot gaits are shown. The gaits are learned using REPS, initialized with the corresponding non-diagonal covariance matrices, specific to each gait. It can be seen that the learned gaits perform according to expected symmetry properties. In Figure~\ref{walk}, one cycle of the walk gait is demonstrated. The legs lift individually in the order front right, hind left, front left, and hind right. In Figure~\ref{trot}, one cycle of the trot gait is displayed. The front left and hind right leg lift up synchronously, as well as the font right and hind left leg. In between a standing posture can be observed.

\subsection{Concrete Initializations}

In this section we discuss how to create a template covariance matrix $\boldsymbol{\Sigma} = \boldsymbol{I} + \boldsymbol{O}$ given the phase gaps specific to a desired gait. Note that this template only indicates the non-zero elements of the scaled covariance matrix defined in (\ref{eq:split_cov}), excluding the parts corresponding to the temporal scaling $\delta_z$ which is always diagonal.
We assume a maximal coupling of $\gamma = 1$, reducing the exploration to a subspace of $\boldsymbol{\Theta}$. In this case the coupling of the movement is identical to the coupling of the exploration, as no parameters outside of the subspace are sampled.

\begin{figure}[H]
  \centering
  \subfloat[Walk covariance initialization]{\includegraphics[width=0.49\textwidth]{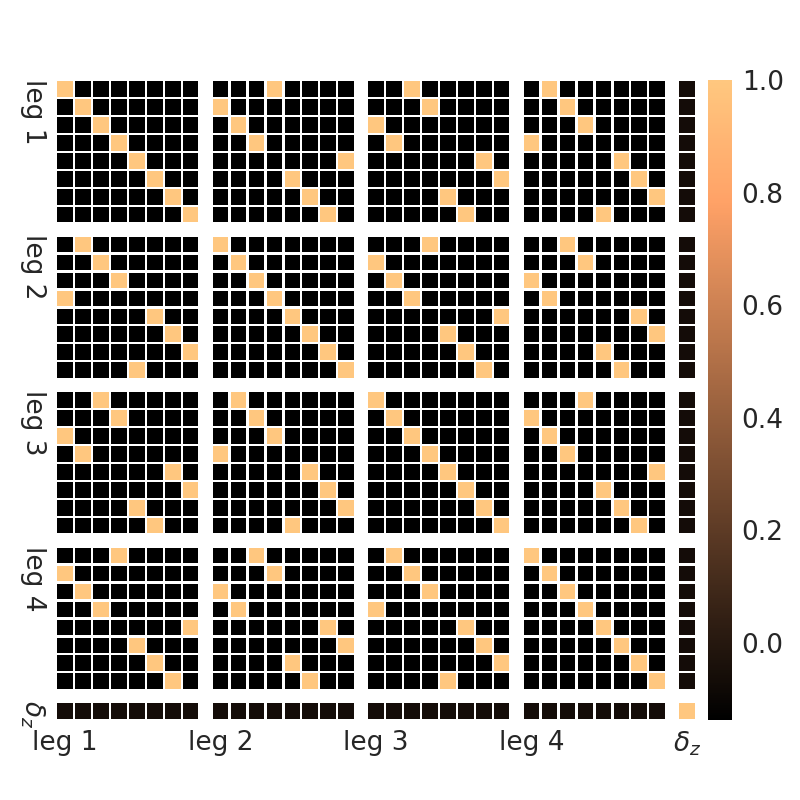}\label{walk_first_cov}}
  \hfill
  \subfloat[Trot covariance initialization]{\includegraphics[width=0.49\textwidth]{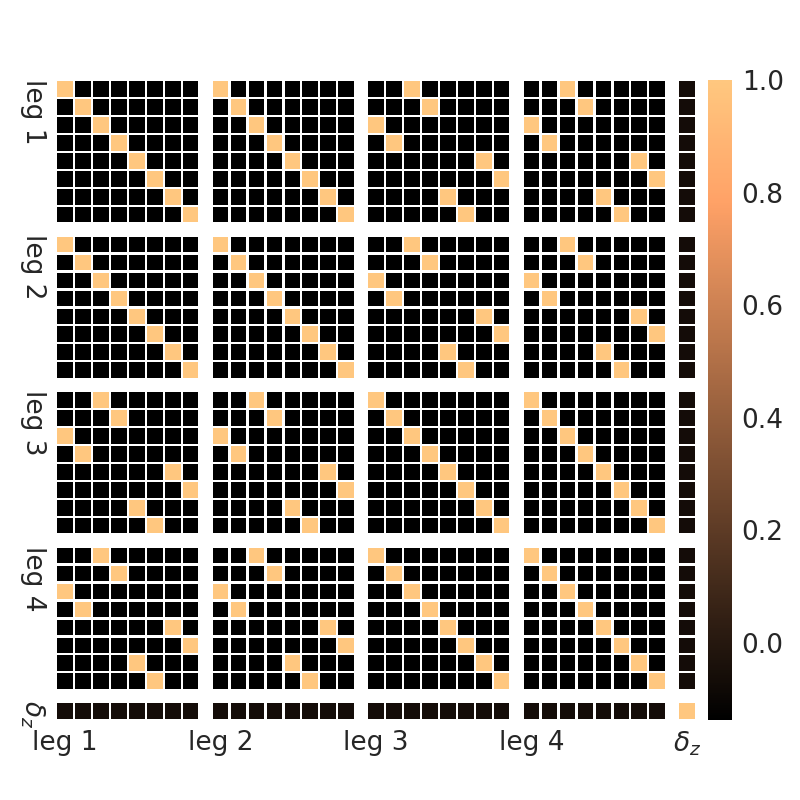}\label{trot_first_cov}}
  \caption{Initial covariance matrices displayed as heatmap.}
  \label{cov_init}
\end{figure}

In Figure \ref{cov_init} the initializations of the covariance matrix $\boldsymbol{\Sigma}$ is displayed as a heat map, divided into 8x8 blocks, corresponding to the coupling of the 8 weights describing one leg trajectory to another. Each of the blocks can further be divided into 4x4 sub-blocks corresponding to the joint-wise coupling. In our setup the knees are not coupled with the hips, therefore only the 4x4 sub-blocks on the diagonal of each 8x8 block are non-zero. 
Using 4 identically distributed cyclic basis functions per joint, any multiple of quarter-phase gaps, can be expressed as a permutation of weights. All 4x4 sub-blocks are identical to the permutation matrix that applies the desired multiple of quarter-phase gaps, which in our case are shifts of the weights. 
Since the phase gaps of the knee trajectories and hip trajectories are the same for each pair of legs, all 8x8 blocks consist of by two identical non-zero 4x4 sub-blocks. Thus, as only four different quarter-phase gaps can be defined, there can only be four different 8x8 blocks.

\vspace{-5mm}

\begin{figure}[H]
  \centering
  \subfloat[Walk covariance matrix after the 3rd update]{\includegraphics[width=0.49\textwidth]{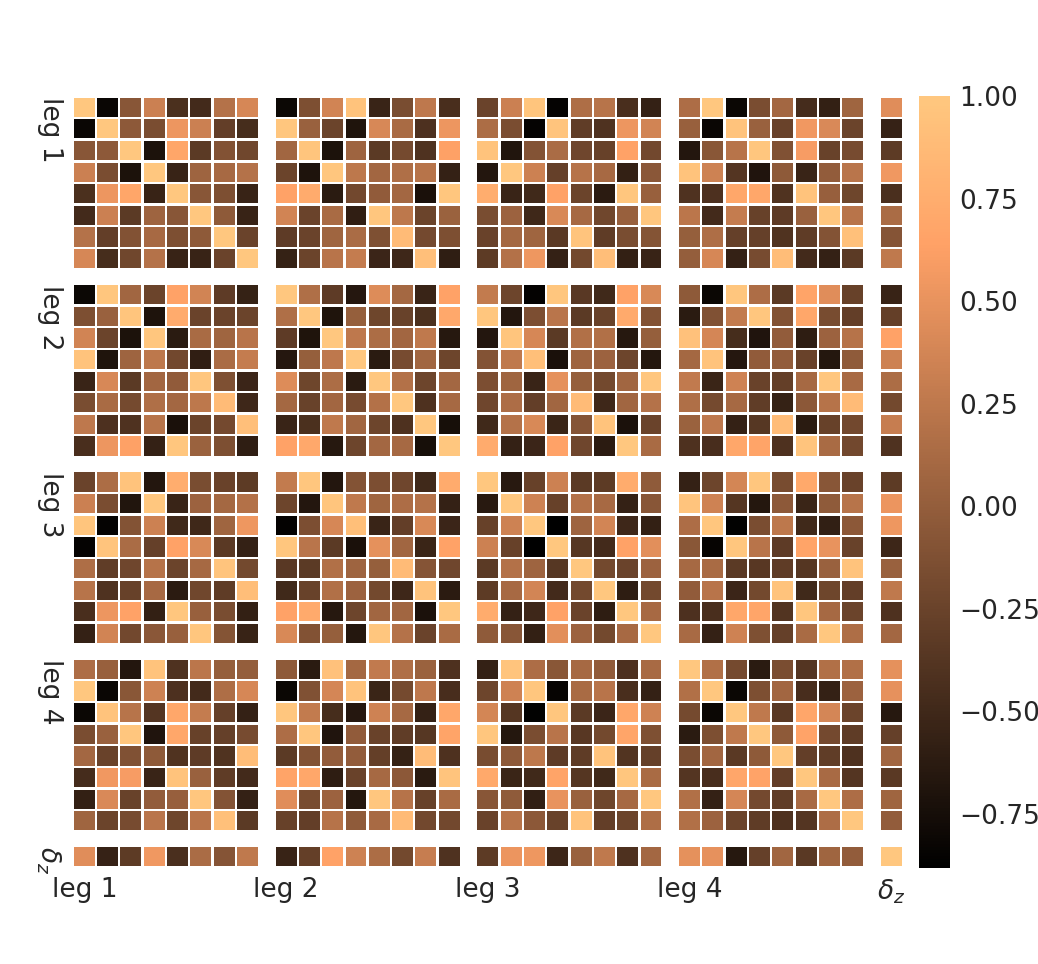}\label{walk_3_cov}}
  \hfill
  \subfloat[Walk covariance matrix after the 4th update]{\includegraphics[width=0.49\textwidth]{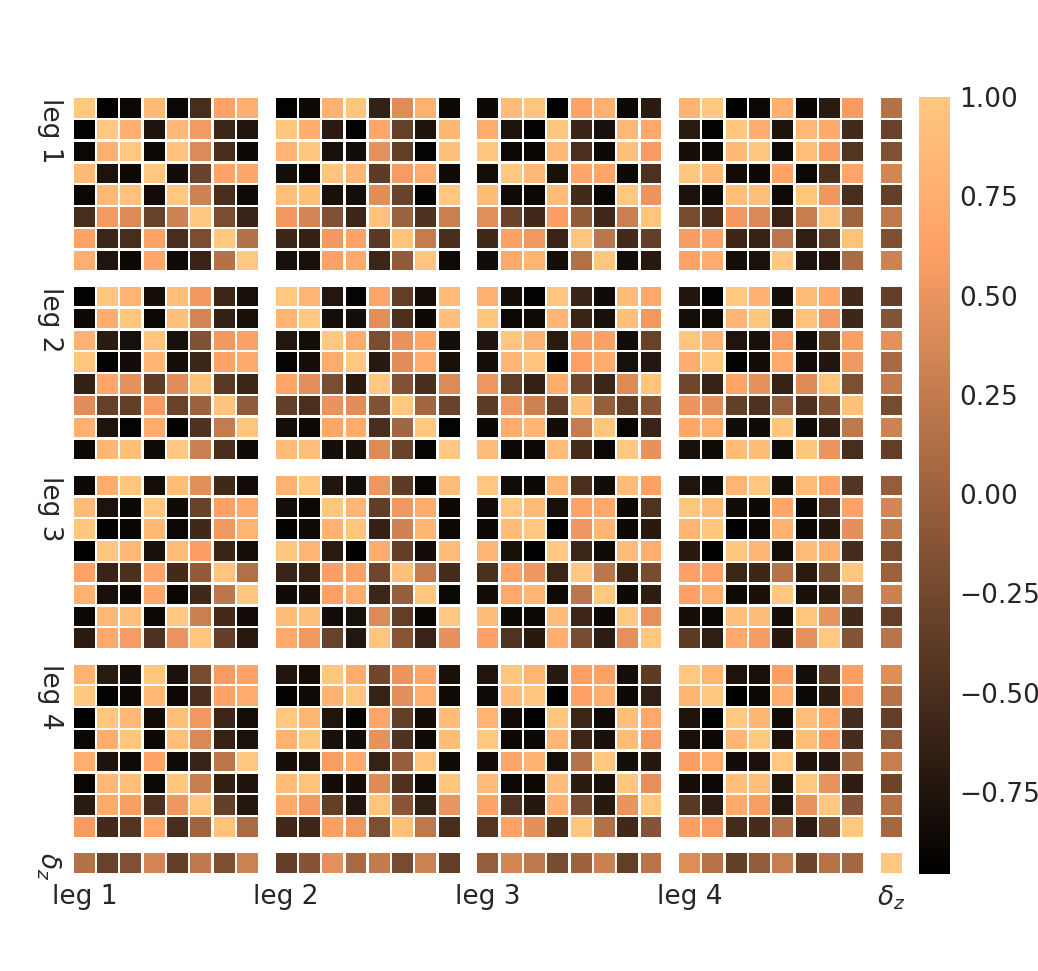}\label{walk_4_cov}}
  \caption{Evolution of the covariance during the learning process of the walk gait with Relative Entropy Policy Search.}
  \label{fig:cov_change}
\end{figure}

In Figure~\ref{trot_first_cov}, it can be observed that leg 1 and leg 2 are coupled to move synchronously with a half-phase gap to leg 3 and leg 4. Whereas in Figure~\ref{walk_first_cov}, all four different 8x8 blocks can be observed, as well as the individual leg movements being delayed by quarter-phase gaps.

\subsection{Change of Exploration in REPS}
Since each iteration of REPS considerably changes the covariance of the search distribution, it diverges from its initialization within a few policy updates. As seen in Figure~\ref{walk_3_cov}, up until the third update, the structure of the initialization for walk gait is preserved, however as shown in Figure~\ref{walk_4_cov}, the structure is lost at the fourth update. Nevertheless, the custom exploration during the first few policy evaluations would still strongly impact the following learning process.

\section{CONCLUSION}
In this paper we presented a new approach to introduce prior knowledge in the context of policy search reinforcement learning, using the policy representation of von Mises basis functions. Namely we initialized the covariance of a Gaussian search distribution, and applied it to the problem of quadruped gait learning. 
We showed that, using REPS, significant improvements in performance can be gained by learning specific types of gaits, i.e., walk and trot compared to random gaits initialized by a diagonal covariance matrix. The different types of gaits can be learned sharing the same linear basis function representation. Further more, REPS has shown to have a more stable learning process, albeit converging prematurely, whereas LRPG shows a slower and unstable learning process.

\section{APPENDIX}

\subsection{Derivation of the Likelihood Ratio Policy Gradient}
\label{apdx:PG}

The expectation is approximated by the sampling. The individual trajectories $\boldsymbol{\tau}_i$ are sufficiently described by the policies parameters $\boldsymbol{\theta}_i$ for an open-loop control. To reduce variance, the mean reward of each batch $\bar{R}$ is used as baseline.
\begin{align*}
	\nabla_{\boldsymbol{\mu}} J_{\boldsymbol{\mu},\boldsymbol{\Sigma}} &= \mathbb{E}_{\boldsymbol{\tau}}\{\nabla_{\boldsymbol{\mu}}\log P(\boldsymbol{\tau}|\boldsymbol{\mu},\boldsymbol{\Sigma})r(\boldsymbol{\tau})\}\\
    &\approx\frac{1}{N}\sum^N_i \nabla_{{\boldsymbol{\mu}}} 
    \log P(\boldsymbol{\tau}_i|{\boldsymbol{\mu}},\boldsymbol{\Sigma})r(\boldsymbol{\tau}_i)\\
    &=\frac{1}{N}\sum^N_i \nabla_{{\boldsymbol{\mu}}} 
    \log\mathcal{N}(\boldsymbol{\theta}_i|{\boldsymbol{\mu},\boldsymbol{\Sigma}})r(\boldsymbol{\theta}_i)\\
    &=\frac{1}{N}\sum^N_i \nabla_{{\boldsymbol{\mu}}} (C-\frac{1}{2}(\boldsymbol{\theta}_i-{\boldsymbol{\mu}})^T\boldsymbol{\Sigma}^{-1}(\boldsymbol{\theta}_i-{\boldsymbol{\mu}}))r(\boldsymbol{\theta}_i)\\
    &=\frac{1}{N}\sum^N_i \boldsymbol{\Sigma}^{-1}(\boldsymbol{\theta}_i-{\boldsymbol{\mu}})r(\boldsymbol{\theta}_i)
\end{align*}
\begin{equation*}
	\nabla_{{\boldsymbol{\mu}}} J_{{\boldsymbol{\mu}},\boldsymbol{\Sigma}} \approx \frac{1}{N}\sum^N_i \boldsymbol{\Sigma}^{-1}(\boldsymbol{\theta}_i-{\boldsymbol{\mu}})(r(\boldsymbol{\theta}_i)-\bar{r})
\end{equation*}

\subsection{Estimation of covariance templates by sampling}

The sub matrix corresponding to the weights in linear function representation is estimated by sampling. Three permutation matrices are devised: $\boldsymbol{P}^{[2]}_1$, $\boldsymbol{P}_1^{[3]}$, and $\boldsymbol{P}_1^{[4]}$, which define the coupling of the weight vectors $\boldsymbol{w}_i$ assigned to each leg w.r.t. leg 1,
\begin{equation*}
\boldsymbol{w}_i = \boldsymbol{P}^{[i]}_1\boldsymbol{w}_1
\end{equation*}
When sampling the weights of the first leg from a gaussian distribution $\boldsymbol{w}_{1,i} \sim  \mathcal{N}(\boldsymbol{0},\boldsymbol{I})$ with zero mean and identity covariance the covariance template can be calculated as
\begin{equation*}
\boldsymbol{\Sigma} = \boldsymbol{I} + \boldsymbol{O} =  \lim_{N\to \infty} \frac{1}{N}  \sum_i^N\boldsymbol{w}_{\mathrm{all},i}^T\boldsymbol{w}_{\mathrm{all},i}, 
\end{equation*}
with $\boldsymbol{w}_{\mathrm{all},i} = (\boldsymbol{w}_{1,i},\boldsymbol{w}_{2,i},\boldsymbol{w}_{3,i},\boldsymbol{w}_{4,i})^T$. The limit can be calculated for sufficiently large $N$  by rounding.


\end{document}